# Conversational implicatures in English dialogue: Annotated dataset

## Elizabeth Jasmi George*, Radhika Mamidi

*LTRC, International Institute of Information Technology, Gachibowli, Hyderabad 500 032, Telangana, India*

---

**Abstract**

Human dialogue often contains utterances having meanings entirely different from the sentences used and are clearly understood by the interlocutors. But in human-computer interactions, the machine fails to understand the implicated meaning unless it is trained with a dataset containing the implicated meaning of an utterance along with the utterance and the context in which it is uttered. In linguistic terms, conversational implicatures are the meanings of the speaker's utterance that are not part of what is explicitly said. In this paper, we introduce a dataset of dialogue snippets with three constituents, which are the context, the utterance, and the implicated meanings. These implicated meanings are the conversational implicatures. The utterances are collected by transcribing from listening comprehension sections of English tests like TOEFL (Test of English as a Foreign Language) as well as scraping dialogues from movie scripts available on IMSDb (Internet Movie Script Database). The utterances are manually annotated with implicatures.

---

---

## 1. Introduction

A conversation in the animation movie 'Anastasia' goes like this,

(1)      *ANYA*: Is this where I get travelling papers?
         *CLERK*: It would be if we let you travel, which we don't so it isn't.

For a simple polar question asked by the character *ANYA*, the complex response by the *CLERK* allows a human reader to conclude on many implicatures with a '*yes*' or a '*no*' as answers to that polar question while reading that dialogue in the movie script. But in human-computer interactions, such a human response will be complicated for the machine to conclude as a '*yes*' or a '*no*'. The basic assumption in a conversation is that, unless otherwise indicated, the participants are adhering to the cooperative principles and maxims [11]. Conversational implicature is the linguistic term for conveying more than what is said, and it is a highly contextualised form of language use that has a lot in common with non-linguistic behaviour [15]. Conversational implicatures are cancellable, non-conventional, calculable and non-detachable. An inference is cancellable or more exactly defensible if it is possible to cancel it by adding some additional premises to the original ones. Implicatures are not part of the conventional meaning of linguistic expressions [6]. As implicatures are not explicitly said, the speaker can always deny an implicature claiming that he/she did not intend to implicate something. As Grice [17] puts it, concerning utterances that carry conversational implicatures, "it is not possible to find another way of saying the same thing, which simply lacks the implicature in question". If an utterance of $P$ conversationally implicates $q$ in $C$, then an utterance of $Q$ conversationally implicates $q$ in $C$, too, given that utterances of $P$ in $C$ and of $Q$ in $C$ say the same thing. This is non-detachability test [2]. Conversational implicature is denoted by the symbol '+>'. For example, in the following utterance by the *Girl* as a response to the *Boy*'s utterance, the implicated meaning is given after the '+>' symbol.

---

* Corresponding author. Tel.:+91 7997245522 .
  *E-mail address:* elizabeth.george@research.iiit.ac.in

(2)     *Boy*: Have you done the trigonometry and calculus problems?
        *Girl*: I did the trigonometry problems.
        +> I did not do the calculus problems.

We required data for training the model in our experiments on computing implicated meanings of utterances. Since datasets with implicatures of utterances are not available, except for a few introductory [3] and sentence level [10] experiments, we attempted to create a dataset of dialogues with their implicature. This paper outlines the sources and methods of data collection for implicatures.

Table 1. Types of Implicatures

| Types of Implicature | Context | Utterance | Implicature |
|---|---|---|---|
| Scalar | Who made these donuts? | I made some of these donuts. | I did not make all of these donuts. |
| Generalised | Did you call John and Benjamin? | I called Benjamin. | I did not call John. |
| Particularised | Did you drink the milk I kept on the table? | The cat seems to be happy. | No. I did not drink milk. The cat might have drunk the milk. |
| Relevance | How about going for a walk? | Isn't it raining out? | No. I am not coming for a walk now. |
| PopeQ (Pope Question) as a response | Are you sure you can take care of yourself this weekend? | Can a duck swim, mother? | Yes. I am sure I can take care of myself. |
| Metaphor as a response | Do you like her? | She is like cream in my coffee. | Yes. I like her a lot. |
| Tautology as a response | Do you want to taste my Hamburger? | Hamburger is hamburger. | No.Hamburgers are not too good to taste. |
| Hyperbole as a response | Are you hungry? | I could eat a horse. | Yes. I am extremely hungry. |
| Idiom as a response | I could have been more careful. | It is useless to cry over spilled milk. | It is useless to be sad about what had already happened. |

*1.1. Generating conversational implicatures*

Conversational implicatures are generated by a variety of situations like replying with a metaphor, idiom, irony, tautology, hyperbole, sarcasm, indirect criticism, etc. PopeQ implicature is where a popularly well-known question is asked in response to a polar question to implicate that the same answer as that of the response question is the answer to the polar question initially asked. The name PopeQ is derived from the question "*Is the Pope Catholic*?" for which the answer is an obvious '*Yes*'. There are classifications of conversational implicatures like scalar, generalised, particularised, relevance-based, etc. See examples in Table 1 where context refers to the most recent text in a conversational context preceding an utterance. The same response can have different meanings in different contexts. For example [16], the response utterance by *B*, "I've cleared the table," has two different implicated meanings for the two different context questions asked by *A*.

(3)     *A*: Have you cleared the table and washed the dishes?
        *B*: I've cleared the table.
        +> I have not washed the dishes.

(4)     *A*: Am I in time for supper?
        *B*: I've cleared the table.
        +> No. You are late for supper.

*1.2. The literature on conversational implicatures*

Conversational implicatures are discussed in detail by Levinson [6] and explained further by Potts [3, 21] reviewing the basic Gricean theory of conversational implicature [17], important consequences, known problems, and useful extensions and modifications. Benotti and Blackburn [15] view conversational implicature as a way of negotiating the meaning in conversational contexts and conveys that context and conversational implicature are highly intertwined. A series of papers by Bouton [18, 19, 20] explores the paradigm of implicatures in pragmatics.

## 2. Approaches attempted for creating implicature corpus

We created a dialogue implicature dataset for which dialogues were collected by transcribing from listening comprehension sections of English language proficiency tests and dialogues from movie scripts. We then annotated them manually with the conversational implicatures to aid our research on generating conversational implicatures in human-computer interactions. A similar annotated resource is an experimental dataset [3] annotated with a definite/ probable '*yes*' or '*no*' for 215 indirect polar questions. For ease and uniformity of creation and usage, we intended to create a dataset with dialogue triplets < *context, utterance, implicature* >, where only one turn of dialogues in a scene for each pair of interlocutors is extracted, and only the single immediate context of an utterance is considered. The number of implicatures for an utterance in a context was not restricted to an upper limit.

*2.1. Related work*

Potts [3] introduced an experimental dataset involving 215 indirect question-answer pairs collected from 4 different sources and annotated with polarity using Amazon Mechanical Turk [1]. Lahiri [10] annotated a corpus of 7,032 sentences using Amazon Mechanical Turk with ratings of formality, informativeness, and implicature for each sentence on a 1-7 scale during which the annotators were asked to form implicatures for a given sentence. Lasecki, Kamar and Bohus [5] collected conversations focused around definable tasks using crowdsourcing methods. In their work, two annotators were assigned part of an agent and a user to a randomly given task and they were asked to engage in a conversation to complete the task. Reddy, Chen and Manning [9] introduced CoQA, a 127k question-answer dataset for building conversational question answering systems. In the above research, pairs of annotators were given passage for reading and were asked to frame questions based on the passage and answer them from the passage consecutively. Another crowd-powered system utilising asynchronous chat for efficient collection of dialogue dataset was designed by Ikeda and Hoashi [4]. In their work, they collected data by giving a topic and asking contributors assigned with part of *A* or *B* to chat upon the given topic for up to 16 turns. Multiple contributors were taking the role of *A* and *B*, and the chat data was not collected in real-time, but instead completed when the contributors were available.

*2.2.Challenges in crowdsourcing the implicature generation task*

On the same lines of the related works mentioned above, we designed a methodology using crowdsourcing platforms to cooperate pairs of contributors and assign them to generate an implicature as follows. A situation will be given to crowdsource-contributors *A* and *B*. Many contributors will be sequentially assigned parts of *A* and *B* for the same given situation so that response utterances in different contexts and their implicatures can be collected. The contributor joining first to attempt the task will be assigned the role of *A,* and another contributor who joins at a later point of time will be assigned the role of *B*. Crowdsource-contributor *A* will be asked to give a single utterance as a polar question based on the given situation. Crowdsource-contributor *B* will be asked to give two utterances as answers to the polar question posted by *A.* (i) An answer without an explicit '*Yes/No*' and (ii) the real implicated meaning of the answer (i) with an explicit '*Yes*' or an explicit '*No*'.

The responding crowdsource-contributors were asked to be cooperative and to give an answer relevant to the question. Some ideal expectations about the question and response in a situation are given in the following Table 2.

Table 2. Examples of situations given for conversation generation task with ideal responses

| Situation | Polar Question | Indirect Answer | Implicated Meaning |
|---|---|---|---|
| Situation 1: The TV had been on for a long time | Can I switch off the tv? | My favourite program will begin now. | No. I am going to watch it now. |
| Situation 1: The TV had been on for a long time | Is anyone watching this? | Oops! I forgot to switch it off. | No. You can switch it off. |
| Situation 2: Both of you, *A* and *B,* are dressed up to go out. | Should I take the umbrella? | It rained yesterday. | Yes. There is a chance of raining. |
| Situation 2: Both of you, *A* and *B,* are dressed up to go out. | Should I take the umbrella? | The sky is black as ink. | Yes. There is a chance of raining. |
| Situation 2: Both of you, *A* and *B,* are dressed up to go out. | Should I take the umbrella? | The sky is clear. | No. It will not rain. |
| Situation 2: Both of you, *A* and *B,* are dressed up to go out. | Should I take the umbrella? | I heard thunder. | Yes. There is a chance of thunderstorms and rain. |

The crowdsourcing platform MicroWorkers' [8] Questions and answers TTV (Template Test & Verification) customised for the implicature annotation task is given in Fig 1. Crowdsource-contributors who are native speakers of English, hailing from the US, UK, Australia, New Zealand and Canada were given this task. This task which demands three inputs requires approximately 8-15 minutes to complete. As the inputs are obtained from 2 different contributors, it takes a turn around time more than 15 minutes. Each utterance was paid with $1.50 on an average, higher than the regular payment of $1.10 for '*conversation generation from outline*' tasks due to the advanced cognitive effort demanded. The crowdsourcing approach did obtain in high quality dialogue data. The challenges in crowdsourcing this kind of dialogue requirements is discussed below.

Fig. 1. MicroWorkers' customised TTV questions and answers template for conversation generation

The first challenge in this approach is gathering and defining the *situations* which can give rise to implicatures in a conversation associated with them. The situation imagined by the task creators may not be entirely comprehended by the crowdsource-contributors. The reading and understanding of the provided *situations* require much cognitive effort and comprehension capabilities from the crowdsource-contributors. Crowdsourcing tasks are generally easy and do not require any specific knowledge and can be annotated directly without much thinking. Since this task demands much imagining from the contributors, it can become a less popular task to attempt and can cost more to get contributors. They might also lose the natural dialogue flow in the process of comprehending the situation.

The inherent chance of the contributors giving irrelevant or indifferent answers like, '*I don't know*', '*I am not sure*' or '*do as you wish*', etc., despite the guidelines given creates another challenge and a requirement upon the completion of the task, to verify each utterance. The quality of the dialogue snippet is much dependent on the question asked by the first crowdsource-contributor who attempted the task for a given *situation*. The difference in the understanding of the *situation* by the two participants *A* and *B* could compromise the quality of the generated utterances. All the questions from first set of contributors that do not give rise to utterances from *B* with a possible implicature have to be deleted from the dataset, and that posed another challenge of cleaning up the data. Crowdsourcing platforms do not follow the FIFO (First In First Out) strategy for tasks, and so the tasks which are not completed quickly or those masked by filters have a high chance of getting forgotten.

The idea that there are clear indicators of implicatures in conversation when metaphors, ironies, indirect criticism, etc. are used, made us think of the possibilities of getting the dialogues from the movie scripts which are clean transcripts of hypothetical human dialogues and the listening comprehension sessions which are designed to create implicated meanings of utterances to test the English language proficiency of non-native speakers of English. Movie scripts had been a source of data for language research particularly for identifying dialogue structures [30], speaker identification [31] and character modelling.

## 3. Implicature corpus

The dialogues are collected from listening comprehension tapescripts of short conversation narrations available online for TOEFL, movie dialogues from the IMSDb [12] for 45 animation movies and other dialogues with metaphors, idioms, hyperboles, indirect criticisms, etc., that are extracted from the internet. Dialogues are also synthesised similar to the extracted ones with an interpretation of answers to polar questions that do not directly express a '*yes*' or '*no*' answer. We selected the animation genre of movies, considering the light tone of the scripts in this genre and the simplicity of the dialogues, as they target children as their primary audience. The movie script data have another advantage of being less noisy and devoid of spelling and grammatical errors compared to the real-time dialogues. The occurrences similar to the ones focused on by de Marneffe [7], involving scalar modifiers such as (5) and numerical answers such as (6) were also identified for the contexts with response utterances obtained from the movie scripts and listening comprehension questions. Online resources are used for the interpretation of idioms, metaphors, hyperboles and tautologies.

(5)     *A: Was the movie wonderful*
              *B: It was worth seeing.*

(6)     *A: Are your kids little?*
              *B: I have a 10-year-old and a 7-year-old.*

### 3.1. Collecting the dialogue snippets

The dialogue snippets with a context and an utterance are identified as a sentence ending with a question mark and not containing multiple question marks in it together with the response sentence that follows it. After scraping those snippets from the IMSDb for the animation genre, they are scrutinised manually for being a polar question and the response holding a chance of implicatures. Those snippets with a '*Yup!*', '*Yes*', '*Yep!*', '*Nope!*','*No*', '*Nay*' and similar, in the response are removed as they give a clear '*yes*' or '*no*' answer to the question asked. The remaining snippets are preprocessed by removing the name of the movie character, making the utterance, and replacing it by *A* for the questioner and *B* for respondent. The preprocessed snippets are manually annotated with one or more implicated meanings, that we infer from the response utterance.

The accuracy of the annotated implicatures can be verified by computing the cosine similarity of annotations by different annotators for the same response utterance. Those annotations with high similarity scores can be prioritised for entry to the dataset.

Fig. 2. Implicature context from (a) movie Script of 'Anastasia'; (b) TOEFL transcript [14]

Fig. 2 shows excerpts from the script of 'Anastasia' movie from IMSDb and a transcript of listening comprehension question from TOEFL for which the part-*A* of the test is the narration of a short conversation between two people with a question about the conversation. Narratives of the TOEFL listening comprehension section are manually transcribed from the English Test Store website [13] for 500 dialogue narrations. Implicature generating dialogue situations are collected from 45 movie scripts of the animation genre from IMSDb and cleaned from other dialogues. Total dialogue snippets from movie scripts are around 500. Some implicatures annotated form listening comprehension sessions are given in Table 3, and those from IMSDb are given in Table 4.

### 3.2. Using the collected dialogue snippets

The obtained data [32] can be used for identifying conversational implicatures in dialogues and for synthesising dialogues with implicatures. This is an ongoing data collection project and when the collected data reaches a considerable scale with negative samples included, it can be used for designing a dialogue system utilising utterances and context embedding for dialogue generation [26, 25] incorporating deep learning approaches such as MrRNN [24], [23] and Locality Sensitive Hashing forest [28] with USE, ELMo or BERT embeddings [29].

Table 3. Utterances and contexts collected from TOEFL listening comprehension with implicated meanings

| Context | Utterance | Implicature |
|---|---|---|
| This calculator is not working, right? | I think you got the battery on upside down. | Yes. It is not working because the battery is not correctly positioned. |
| Would you like to go with us for coffee a little later? | I am off caffeine. Medical restriction. | No. I have to eliminate coffee from my diet. |
| Were you pleased with last week's convention? | Nothing went as planned. | No. I was not pleased with last week's convention. |
| Let me help you with those packages? | Thanks, but it is only three quarters to the block. | No. You don't have to help me with those packages. It is not too far for me to carry the packages. |

Table 4. Utterances and context collected from IMSDb movie dialogues with Implicated meanings

| Context | Utterance | Implicature | Movie Name |
|---|---|---|---|
| Can I call you in a little while? | It's four in the morning... I'm going to go to sleep. | No. You should not call me. | Lost in Translation |
| It's bad isn't it? | We should get you to the doctor. | Yes. It is bad. | Lost in Translation |
| And marriage, does that get easier? | It's hard. We started going to a marriage counselor. | No. Marriage did not get easier. | Lost in Translation |
| How does that sound? | About as bad as you smell! | That does not sound good. | Anastasia |

## 4. Conclusion and future work

In this paper, we present our approach to collect and annotate an introductory dataset of dialogues with implicatures associated with the response utterance. The collected dataset [32] can be used as a reference for identifying and synthesising conversational implicatures. The dialogues are collected from 74 listening comprehension short conversation practice sections of the TOEFL English proficiency test and 45 movie scripts of the animation genre. The paper also outlines the challenges faced while trying to crowdsource the dialogue collection and implicature annotation. As implicatures are generated in a wide range of situations and are highly dependent on the hearer's understanding, we have primarily focused on the polar questions where an indirect answer without an explicit 'Yes' or 'No' generates implicatures. In the future, we are planning to add more contexts along with the polar question context considered in this paper and annotate the identified dialogues with implicatures. In our future work, scalar implicatures which can be identified with the comparison keywords [6] such as < all, most, many, some, few >, < always, often, sometimes >, < must, should, may > would be focused in particular using a similarity-Judgement method like that proposed by Degen [27]. Scalar implicatures are easy to isolate and notice and a lot of research on implicature [22] is focused on those implicatures. Extra context features where context/i refers to the ith most recent additional context would also be considered where the utterance gets its meaning from multiple-context-Utterances going back in time during a conversation.

# References

[1] Amazon Mechanical Turk Home page https://www.mturk.com/ accessed 2019/09/19.

[2] Blome-Tillmann, M. (2013) "Conventional implicatures (and how to spot them)", Philosophy Compass, 8, 170–185.

[3] Christopher Potts, "Background on conversational implicature, Stanford Linguistics", LING7800-007: Computational Pragmatics, Retrieved from http://compprag.christopherpotts.net/ accessed 2019/09/19.

[4] Kazushi Ikeda, Keiichiro Hoashi (2018) "Utilizing Crowdsourced Asynchronous Chat for Efficient Collection of Dialogue Dataset", The Sixth AAAI Conference on Human Computation and Crowdsourcing HCOMP

[5] Lasecki, W.S., Kamar, E., & Bohus, D. (2013) "Conversations in the crowd: Collecting data for task oriented dialogue learning". In Proceedings of AAAI.

[6] Levinson, Stephen C. (1983) "Pragmatics". Cambridge: Cambridge University Press, doi: https://doi.org/10.1017/CBO9780511813313

[7] Marie-Catherine de Marneffe (2012) "WHAT'S THAT SUPPOSED TO MEAN? MODELING THE PRAGMATIC MEANING OF UTTERANCES", Phd thesis, Stanford University, Retrieved from https://nlp.stanford.edu/ accessed 2019/09/19.

[8] MicroWorkers Homepage https://www.microworkers.com/ accessed 2019/09/19.

[9] Reddy, S., Chen, D., Manning, C.D. (2019) "CoQA: A Conversational Question Answering Challenge", MIT Press, Transactions of the Association for Computational Linguistics, Volume 7, p.249-266, doi: https://doi.org/10.1162/tacl_a_00266

[10] S. Lahiri (2015) "SQUINKY! A Corpus of Sentence-level Formality, Informativeness, and Implicature", CoRR, vol. abs/1506.02306.

[11] Yule, G. (1996), "Pragmatics", Oxford: Oxford University Press, doi: https://doi.org/10.1017/CBO9780511757754.011

[12] Internet Movie Script Database (IMSDb) Homepage http://www.imsdb.com/ accessed 2019/09/19.

[13] English Test store Home page https://englishteststore.net/ accessed 2019/09/19.

[14] ETS home page https://www.ets.org/toefl accessed 2019/09/19.

[15] Benotti, L., Blackburn, P. (2014) "Context and implicature." In Brezillon, P.,Gonzalez, A. (eds.), Context in Computing. Springer, New York, 419-436.

[16] Mahmood, R. (2015) "A Pragmatic Analysis of Imference as a Collective Term for Implicature and Inference" International Journal on Studies in English Language and Literature, 3(9). Retrieved from; https://www.arcjournals.org/pdfs/ijsell/v3-i9/8.pdf accessed 2019/09/19.

[17] H. Paul Grice. (1975), "Logic and conversation." In Cole, P., and J.L.Morgan, eds. Speech Acts. New York: Academic Press, 41–58, doi: https://doi.org/10.1057/9780230005853_5.

[18] Bouton, L. F. (1988) "A cross-cultural study of the ability to interpret implicatures in English". World Englishes, 7(2), 183-97.

[19] Bouton, L. F. (1989) "So they got the message, but how did they get it?", IDEAL, 4, 119-49.

[20] Bouton, L. F. (1996) "Pragmatics and Language Learning" Monograph Series Volume 7, p1-20.

[21] Christopher Potts. (2012) "Conversational implicature: an overview", Ling 236: Context dependence in language and communication, Spring 2012, Retrieved from https://web.stanford.edu/class/linguist236/implicature/materials/ling236-handout-04-02-implicature.pdf

[22] Eiteljoerge, S.F.V., Pouscoulous, N., Lieven, E.V.M. (2018) "Some pieces are missing: implicature production in children", Front. Psychol. 9 (1928) doi: https://doi.org/10.3389/fpsyg.2018.01928.

[23] Hartshorne, J. K., Snedeker, J., Liem Azar, S. Y.-M., and Kim, A. E. (2015) "The neural computation of scalar implicature", Lang. Cogn. Neurosci. 30, 620–634. doi: 10.1080/23273798.2014.981195.

[24] I. V. Serban, T. Klinger, G. Tesauro, K. Talamadupula, B. Zhou, Y. Bengio, and A. Courville (2016) "Multiresolution Recurrent Neural Networks: An Application to Dialogue Response Generation", arXiv preprint arXiv:1606.00776.

[25] J. D. Williams and G. Zweig (2016) "End-to-end lstm-based dialog control optimized with supervised and reinforcement learning", arXiv preprint arXiv:1606.01269.

[26] Alexander Bartl and Gerasimos Spanakis (2017) "A retrieval-based dialogue system utilizing utterance and context embeddings", CoRR, abs/1710.05780.

[27] Degen, Judith (2015) "Investigating the distribution of some (but not all) implicatures using corpora and web-based methods", Semantics & Pragmatics 8:11:1–55.

[28] M. Bawa, T. Condie, and P. Ganesan (2005) "LSH Forest: Self-Tuning Indexes for Similarity Search. In International Conference on World Wide Web", pages 651–660.

[29] Matthew Henderson, Pawel Budzianowski, Iñigo Casanueva, Sam Coope, Daniela Gerz, Girish Kumar, Nikola Mrkšic, Georgios Spithourakis, Pei-Hao ´ Su, Ivan Vulic, and Tsung-Hsien Wen (2019) "A repository of conversational datasets", In Proceedings of the 1st Workshop on Natural Language Processing for Conversational AI.

[30] Banchs, R. E, "Movie-DiC: A movie dialogue corpus for research and development". In Proceedings of the 50th Annual Meeting of the Association for Computational Linguistics, 2012, 203–207.

[31] Kundu, Amitava, Dipankar Das and Sivaji Bandyopadhyay. "Speaker identification from film dialogues." 2012 4th International Conference on Intelligent Human Computer Interaction (IHCI) (2012): 1-4.

[32] George, Elizabeth Jasmi (2019): Implicature dataset. figshare. Dataset. https://doi.org/10.6084/m9.figshare.10315505.v3 Retrieved from https://figshare.com/articles/Implicature dataset/10315505.